\documentclass{article}
\usepackage{spconf,amsmath,graphicx}
\usepackage{amssymb}
\usepackage{tabularx}
\usepackage{epstopdf}

\title{Salient Object Detection via Objectness Measure}

\twoauthors
  {Sai Srivatsa R }
	{Indian Institute of Technology, Kharagpur}
  {R. Venkatesh Babu}
	{Video Analytics Lab, SERC\\
Indian Institute of Science, Bangalore}
\begin{document}
\ninept
\maketitle
\begin{abstract}

Salient object detection has become an important task in many image processing applications. The existing approaches exploit background prior and contrast prior to attain state of the art results. In this paper, instead of using background cues, we estimate the foreground regions in an image using objectness proposals and utilize it to obtain smooth and accurate saliency maps. We propose a novel saliency measure called `foreground connectivity' which determines how tightly a pixel or a region is connected to the estimated foreground. We use the values assigned by this measure as foreground weights and integrate these in an optimization framework to obtain the final saliency maps. We extensively evaluate the proposed approach on two benchmark databases and demonstrate that the results obtained are better than the existing state of the art approaches.
\end{abstract}
\begin{keywords}
Image Saliency, Objectness Proposals, Image Segmentation, Superpixels
\end{keywords}
\section{Introduction}
\label{sec:intro}

The Human visual system has the ability to process parts of image which are relevant, discarding the rest. This helps us to perceive objects even before identifying them. Saliency detection, i.e. computationally detecting these relevant regions is a complex problem which takes cues from models in cognitive psychology, neurobiology and computer vision. It has gained a lot of attention in the recent years from the computer vision community owing to its use in object recognition \cite{04cvpr/RutishauserWWKP}, object segmentation \cite{09iccv/donoser_saliency}, image re-targeting \cite{imgrt} and cropping \cite{imgcp}, image retrieval \cite{Sketch2Photo} etc. 

Works in saliency detection are classified into three categories : fixation prediction, Salient object detection and Objectness proposal generation. 

Early models were biologically inspired and were evaluated on human eye fixation datasets. Ullman and Koch \cite{85HN/KochVisualAttention} define saliency at a given location as how different it is from its surrounding in color, orientation, motion, depth etc. Itti et al. \cite{98pami/Itti} follow the same framework and propose a centre-surround contrast using Difference of Gaussian to obtain their Saliency maps. Ma and Zhang \cite{03ACMMM/Ma-Contrast-based} use similar contrast analysis and extend it using fuzzy growth model. These models are evaluated on eye fixation databases. These models highlight edges and corners and are not suitable for detecting complete salient regions.

Salient object detection models aim to segment the object as a whole and are evaluated mostly on data labeled by humans such as bounding boxes or Foreground masks. These methods use low level cues such as contrast prior \cite{09cvpr/Achanta_FTSaliency,11cvpr/Cheng_Saliency,10cvpr/goferman_context} or boundary prior \cite{manifoldranking,salopt,geosal}. Methods using contrast prior rely on uniqueness of the object and contrast between pixels or regions, center-surround differences etc. Methods based on contrast prior may be further classified into global or local methods. Local methods involve computing contrast measure in a local patch, whereas global methods use the entire image to compute the saliency. Global methods often use spatial feature as two pixels/regions which look similar but are far away need not belong to the same object. Global methods fail when the background is complex. 
Achanta et al. \cite{09cvpr/Achanta_FTSaliency} computes saliency based on pixels color difference from the mean image color. Cheng. et al \cite{11cvpr/Cheng_Saliency} combines global contrast with spatial differences to generate a saliency map. Perazzi et al. \cite{PerazziKPH12} computes saliency by decomposing the image into homgeneous elements. Contrast and spatial distribution are used to obtain pixel-accuarate saliency maps. These are estimated using high-dimensional Gaussian filter.

However, contrast prior alone is not very effective. The other most commonly used cue is based on the assumption that most photographers do not crop the salient object along the view frame. Hence the image boundary forms the background. However boundary prior is fragile and its prone to fail even when the object is slightly touching the background. Wei et al. \cite{geosal} propose a saliency measure based on shortest length between an image patch and a virtual boundary node and it overcomes the shortcomings of boundary prior by connecting the boundary regions to a virtual node through an edge with suitable boundary weight. Yang et al. \cite{manifoldranking} ranks similarity of image regions with foreground or background cues using a graph-based manifold ranking. The ranking is based on relevance of an element with respect to the given queries. Zhu et al. \cite{salopt} propose a boundary connectivity measure that utilizes both contrast prior and boundary prior. Foreground and Background weights obtained are then combined using an optimization framework.

\begin{figure*}[t]
\label{phases}
\centering
\includegraphics[height=.14\linewidth]{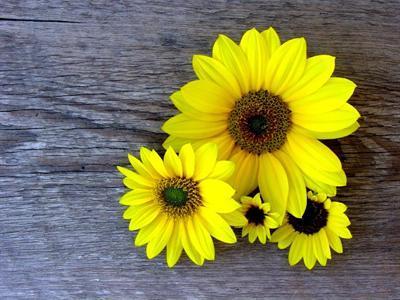}
\includegraphics[height=.14\linewidth]{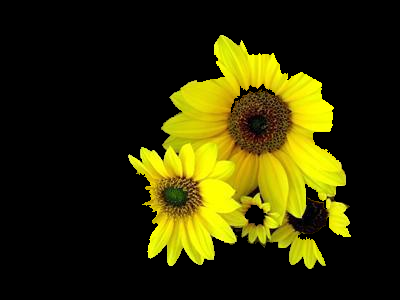}
\includegraphics[height=.14\linewidth]{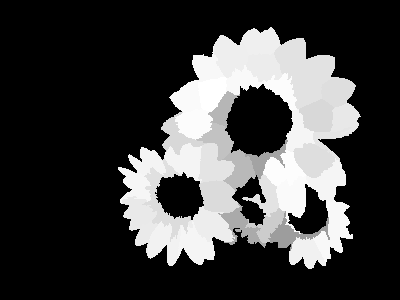}
\includegraphics[height=.14\linewidth]{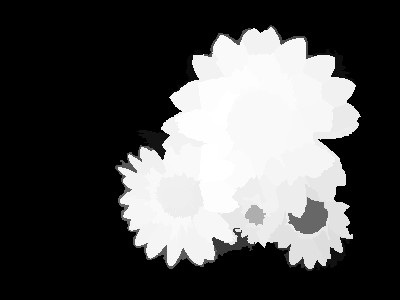}
\includegraphics[height=.14\linewidth]{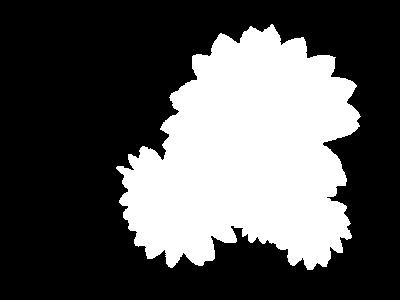}

\centerline{(a)~~~~~~~~~~~~~~~~~~~~~~~~~~~~~~~~~~~~~~(b)~~~~~~~~~~~~~~~~~~~~~~~~~~~~~~~~~~~~~~~(c)~~~~~~~~~~~~~~~~~~~~~~~~~~~~~~~~~~~~~~~(d)~~~~~~~~~~~~~~~~~~~~~~~~~~~~~~~~~~~~~(e)}
\caption{Illustration of main phases of our algorithm. (a) Input Image. (b) Thresholded Objectness Map. (c) Foreground weights (d) Saliency Map after Optimization (e) Ground truth}
\end{figure*}

Objectness proposal generation methods propose small number of windows that are likely to contain the object in an image thereby reducing search space for classifiers. Alexe et al. \cite{alexe2012measuring} propose an objectness measure that combines several image cues measuring an objects' characteristics in a Bayesian framework. Zhang et al. \cite{zhang2011proposal} propose cascaded ranking SVM to generate an ordered set of proposals. Cheng et al. \cite{BingObj2014} proposes a binarized version of normed gradient features (BING) which can be tested using few atomic operations to generate Objectness proposals.

Jiang et al \cite{jiang} integrates Objectness with Uniqueness and Focusness to obtain saliency maps. However these maps are not smooth and it is difficult to attribute these results to specific algorithm properties \cite{PerazziKPH12}

In this work, rather than obtaining the background from image boundary and using the boundary prior, we quickly obtain a rough estimate of foreground regions by utilizing a modified version of the recently proposed objectness proposal technique \cite{BingObj2014}. We then compute super-pixel objectness which is a measure that quantifies how likely it is for a super-pixel to be a part of the foreground. The foreground and background regions are obtained by appropriately thresholding the above measure. We propose a robust saliency measure called foreground connectivity which assigns saliency values to these super-pixels.

Rather than combining the cues heuristically (weighted summation or multiplication), we use a principled optimization framework proposed by \cite{salopt} that regards saliency as a global optimization problem. The values assigned to the super-pixels are is considered as the foreground weights in the cost function. Minimizing the cost function would result in foreground regions taking higher values and background regions taking lower values. The obtained maps are also smooth and uniform due to the smoothness constraint in the cost function. 

We have extensively evaluated our method on MSRA-1000 \cite{09cvpr/Achanta_FTSaliency} and CSSD dataset \cite{Yan:13} and we show that the proposed method performs at par or even better than the existing methods. 

The paper is organized as follows. Section 2 describes various modules of the proposed approach including the new `Foreground connectivity' metric. Experiments and results are discussed in section 3 and sec. 4 concludes the paper. 

\section{Methodology}
The proposed approach is as follows. We build an Objectness Map using Objectness Proposals to capture super-pixels
containing the object. Next, using the foreground connectivity measure, we assign foreground weights to super-pixels. We use Saliency optimization technique to combine our foreground weights with background measure as
used in \cite{salopt} to obtain smooth and accurate saliency maps. Figure 1 illustrates the main phases of our algorithm.

\subsection{Objectness Map}

Objects are stand-alone things with well-defined closed boundaries and centers \cite{alexe2012measuring}. When windows containing objects are resized to a smaller size, the magnitude of norm of image gradients (NG) become good discriminative features. These normed gradients are inert to change of scale/aspect ratio and translation. The fact that objects share some correlation in the NG spaces is utilized in BING \cite{BingObj2014} to detect objects. The image is resized into fixed sizes and the normed gradient values in 8 $\times$ 8 region is used as 64 dimensional normed gradient feature. These windows are scored with a learned linear model $\mathbf{w} \in \mathbb{R}^{64}$.

\begin{equation}\label{equ:score}
s_l = \langle\mathbf{w},\mathbf{g}_l\rangle,
\end{equation}
\begin{equation}\label{equ:location}
l = (i,x,y),
\end{equation}
where $s_l$, $\mathbf{g}_l$, $l$, $i$ and $(x,y)$ are filter score, NG feature, location, size and position of a window respectively. Using non-maximal suppresion, top proposals are chosen and are re-ranked based on their location and size using coefficients learned using a linear SVM. BING is accurate and extremely fast as the features are binarized and only a few atomic operations are required for obtaining the objectness proposals.

In the proposed approach we have adapted BING by modifying the following modules. Instead of using a learned linear model, we use an 8 $\times$ 8 Laplacian of Gaussian like filter and obtained scores for the windows. We skip re-ranking them based on the learned coefficients. The proposed model has higher weights placed along the edges and it resembles the center-surround patterns \cite{85HN/KochVisualAttention}. With this model, we also cut down the training time.

After obtaining the Objectness proposals, we generate the Objectness Map. Objectness score of a window tells us how likely it is to contain an object. We use these objectness proposals to obtain pixel-wise objectness ($PixObj$) score which tells us how likely it is for a pixel to be a part of an object. Pixel-wise objectness score is given by
\begin{equation}\label{equ:Pix-Obj}
PixObj(p) = \sum_{i=1}^{k} s_{i} G_{i} (x,y)
\end{equation}
where $s_{1}, s_{2} ... s_{k}$ are the objectness scores of the proposals containing pixel $p$ and $G_{i}$ is a Gaussian window having same dimensions as that of the given proposal , $x$ and $y$ are relative $x$ and $y$ coordinate of pixel $p$ with respect to the given proposal.

Sum of pixel-wise object probability in a super pixel region gives us Objectness score of that super pixel region (which is used to construct the Objectness Map).
\begin{equation}\label{equ:Objectness}
Objectness(R) = \sum_{ i \in R} PixObj(p_{i})
\end{equation}
where $p_{i}$ is a pixel belonging to super pixel region $R$.
To obtain super pixel regions, we use SLIC \cite{slic} as it is fast and it preserves boundaries.

We use adaptive thresholding to obtain the Objectness Map. We observe that these Objectness Maps in few cases are able to segment out the salient object completely, however most objectness maps either miss out parts of object or include parts of background in it. This happens as the Objectness proposals are rectangular regions containing both foreground and background.

\begin{figure} [h]
\label{objmaps}
\centering
\begin{minipage}{0.23\linewidth}
\includegraphics[width=\linewidth]{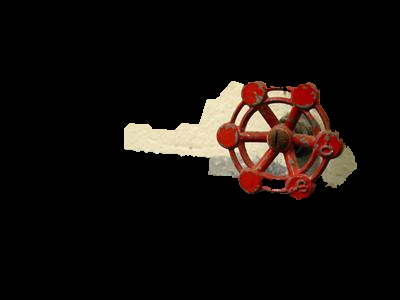}
\end{minipage}
\begin{minipage}{0.23\linewidth}
\includegraphics[width=\linewidth]{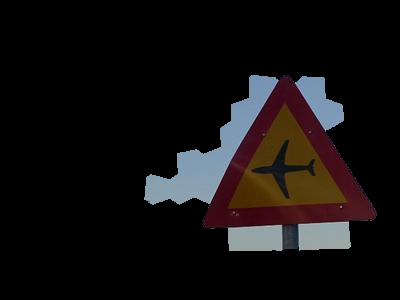}
\end{minipage}
\begin{minipage}{0.23\linewidth}
\includegraphics[width=\linewidth]{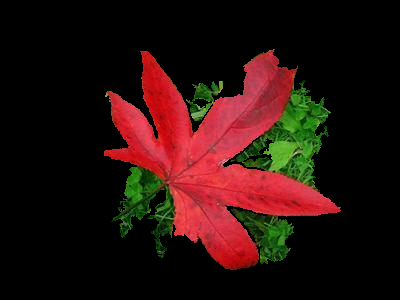}
\end{minipage}
\begin{minipage}{0.23\linewidth}
\includegraphics[width=\linewidth]{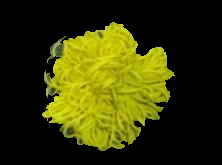}
\end{minipage}

\begin{minipage}{0.23\linewidth}
\includegraphics[width=\linewidth]{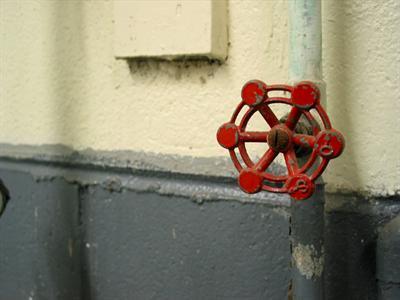}
\end{minipage}
\begin{minipage}{0.23\linewidth}
\includegraphics[width=\linewidth]{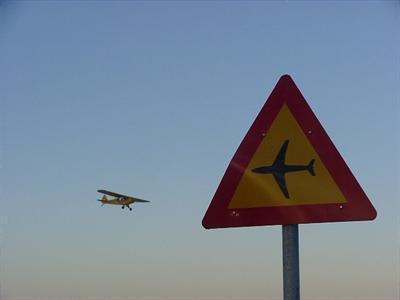}
\end{minipage}
\begin{minipage}{0.23\linewidth}
\includegraphics[width=\linewidth]{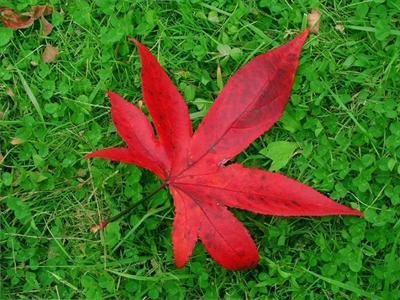}
\end{minipage}
\begin{minipage}{0.23\linewidth}
\includegraphics[width=\linewidth]{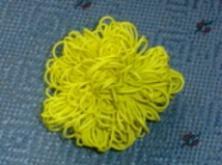}
\end{minipage}
\caption{Rough estimate of foreground obtained using Objectness Maps}
\end{figure}

\begin{figure*} [t]
\label{results}
\centering

\includegraphics[height=.09\linewidth]{Figures/010015}
\includegraphics[height=.09\linewidth]{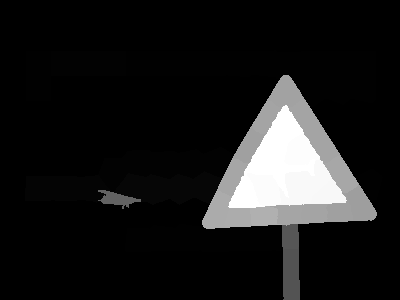}
\includegraphics[height=.09\linewidth]{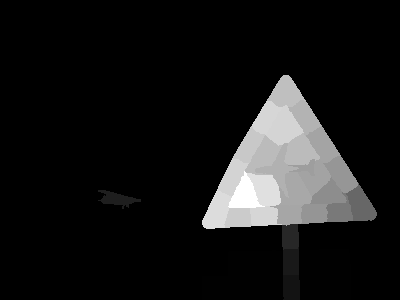}
\includegraphics[height=.09\linewidth]{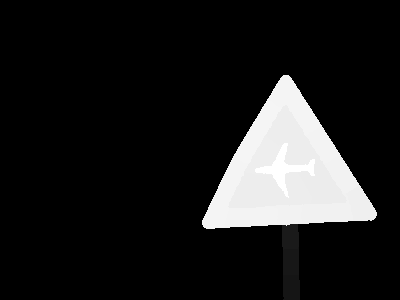}
\includegraphics[height=.09\linewidth]{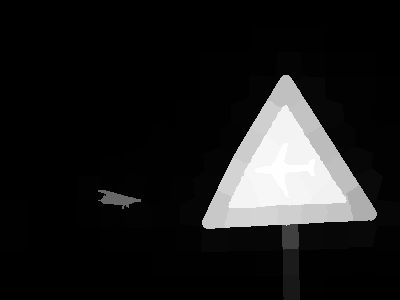}
\includegraphics[height=.09\linewidth]{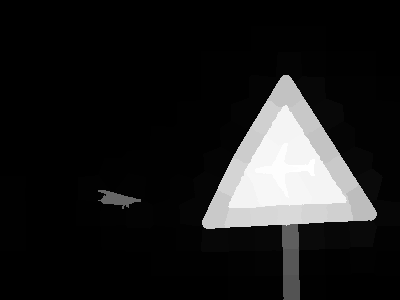}
\includegraphics[height=.09\linewidth]{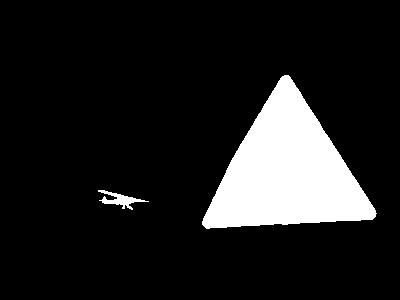}
\\[2pt]
\includegraphics[height=.09\linewidth]{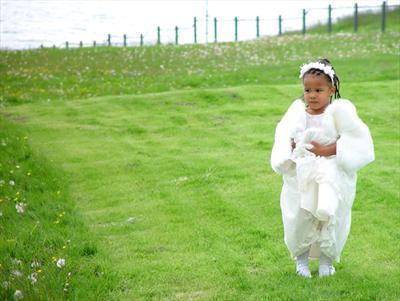}
\includegraphics[height=.09\linewidth]{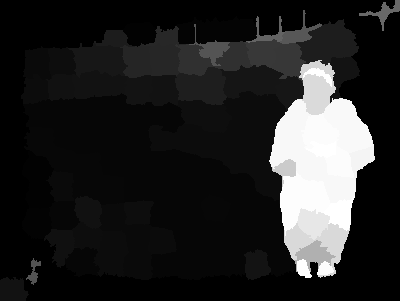}
\includegraphics[height=.09\linewidth]{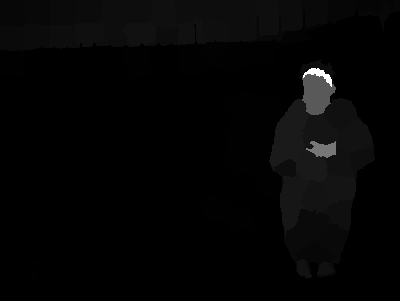}
\includegraphics[height=.09\linewidth]{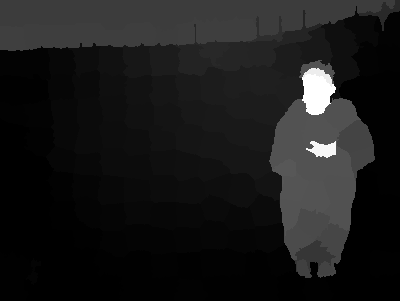} \includegraphics[height=.09\linewidth]{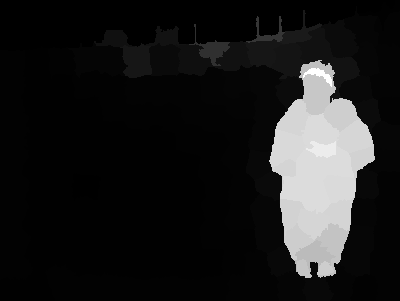}
\includegraphics[height=.09\linewidth]{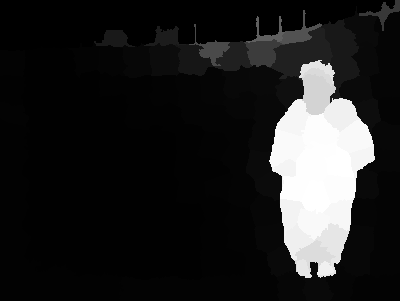}
\includegraphics[height=.09\linewidth]{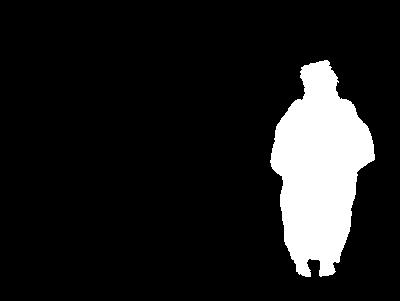}
\\[2pt]
\includegraphics[height=.09\linewidth]{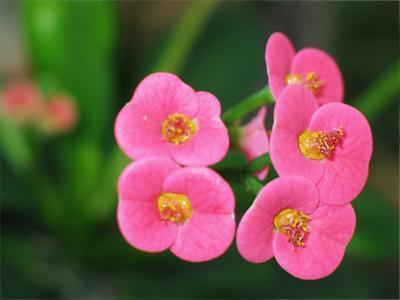}
\includegraphics[height=.09\linewidth]{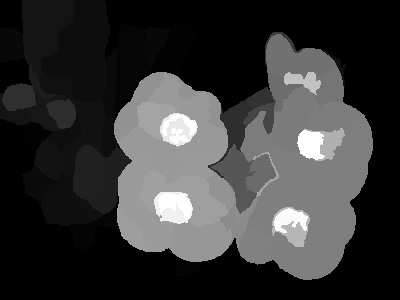} \includegraphics[height=.09\linewidth]{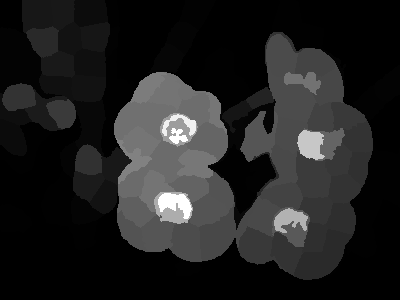}
\includegraphics[height=.09\linewidth]{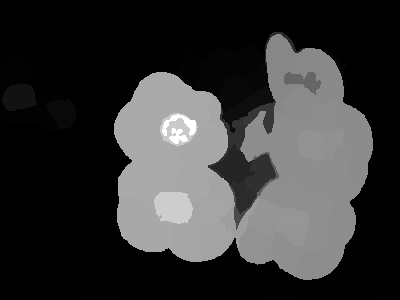} \includegraphics[height=.09\linewidth]{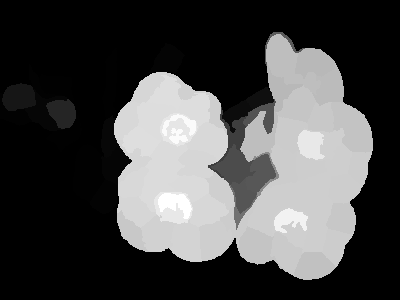} \includegraphics[height=.09\linewidth]{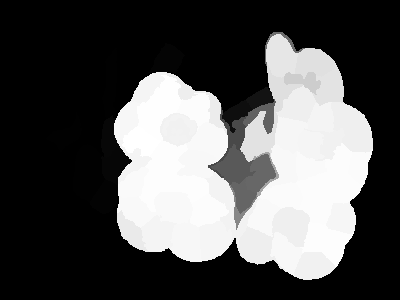}
\includegraphics[height=.09\linewidth]{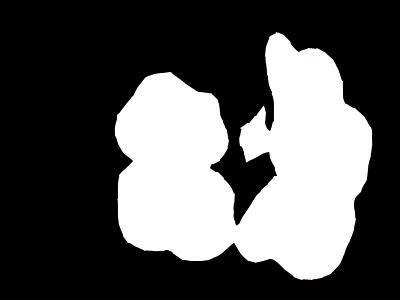}
\\[2pt]
\includegraphics[height=.08\linewidth]{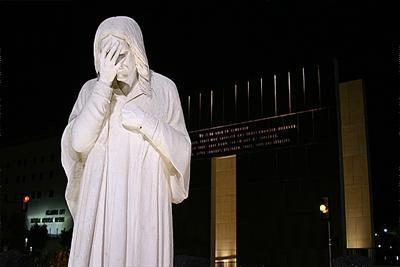}
\includegraphics[height=.08\linewidth]{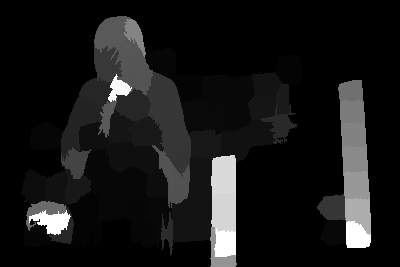} \includegraphics[height=.08\linewidth]{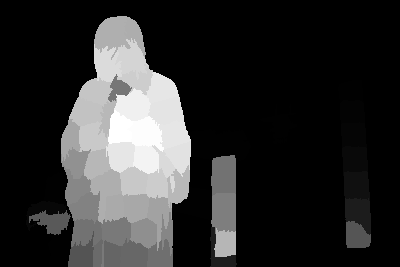}
\includegraphics[height=.08\linewidth]{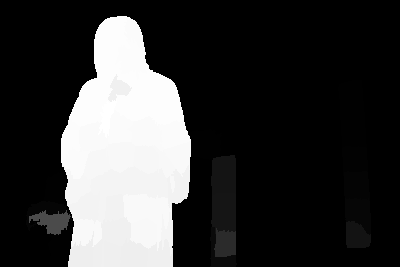} \includegraphics[height=.08\linewidth]{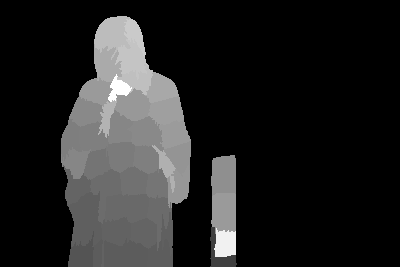} \includegraphics[height=.08\linewidth]{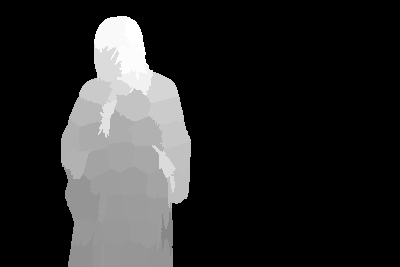}
\includegraphics[height=.08\linewidth]{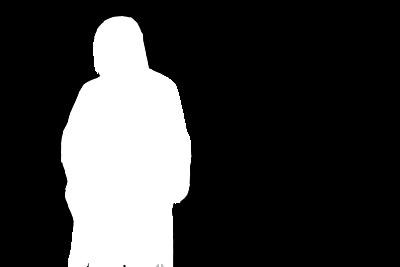}
\\[2pt]
\includegraphics[height=.093\linewidth]{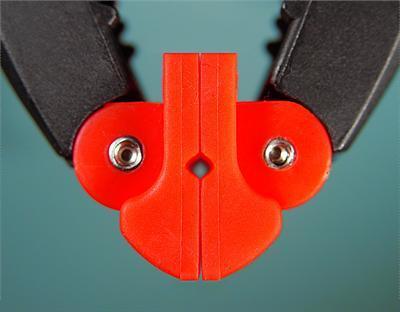}
\includegraphics[height=.093\linewidth]{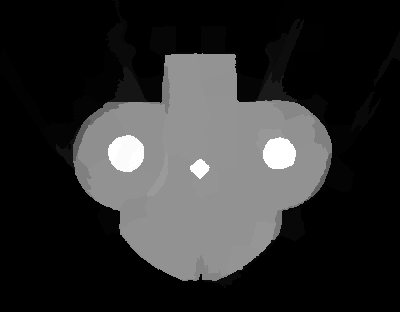} \includegraphics[height=.093\linewidth]{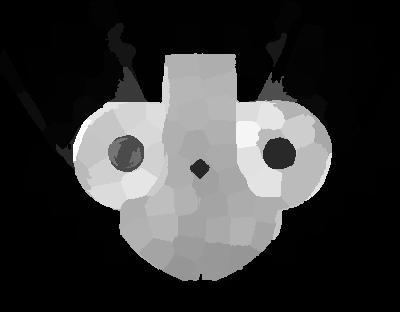}
\includegraphics[height=.093\linewidth]{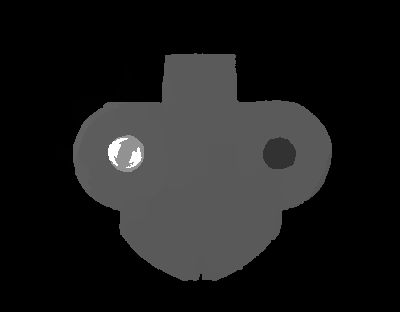} \includegraphics[height=.093\linewidth]{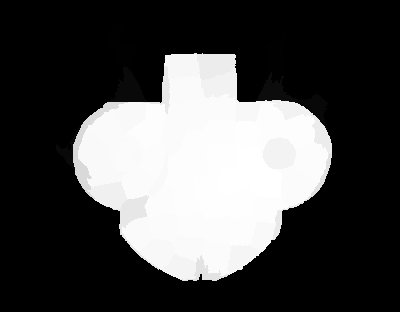} \includegraphics[height=.093\linewidth]{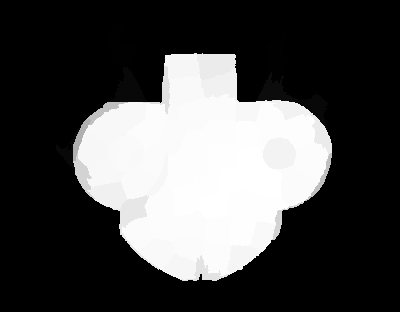}
\includegraphics[height=.093\linewidth]{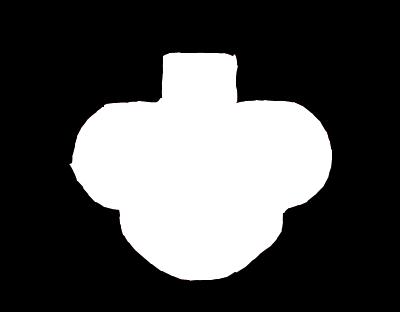}
\\[4pt]
\includegraphics[height=.08\linewidth]{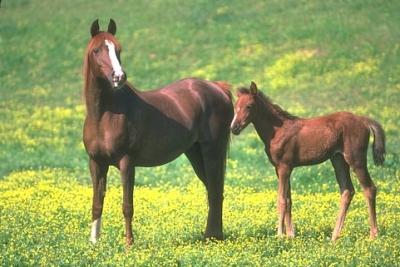}
\includegraphics[height=.08\linewidth]{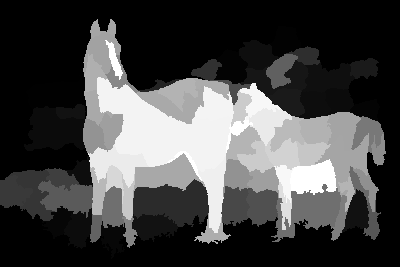} \includegraphics[height=.08\linewidth]{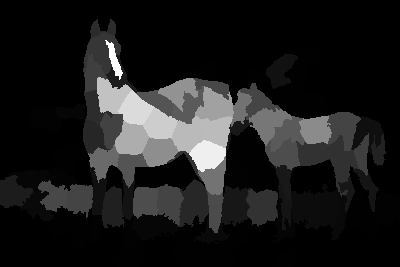}
\includegraphics[height=.08\linewidth]{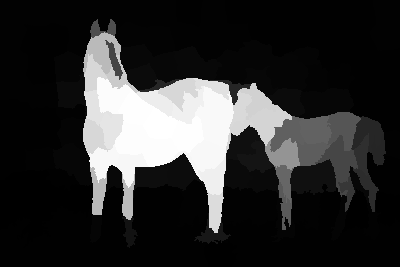} \includegraphics[height=.08\linewidth]{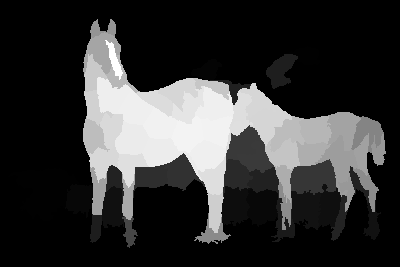} \includegraphics[height=.08\linewidth]{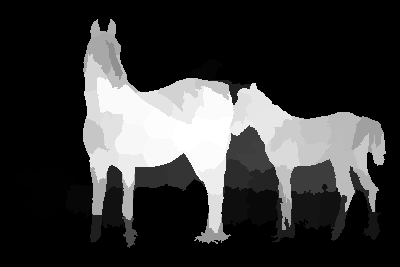}
\includegraphics[height=.08\linewidth]{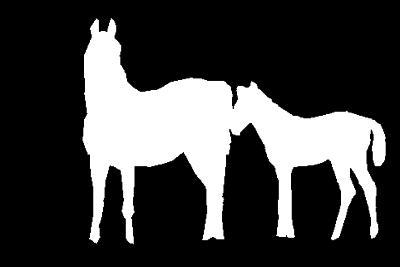}
\\[2pt]
\includegraphics[height=.08\linewidth]{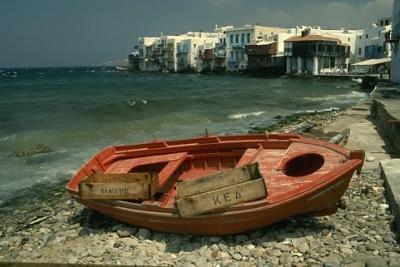}
\includegraphics[height=.08\linewidth]{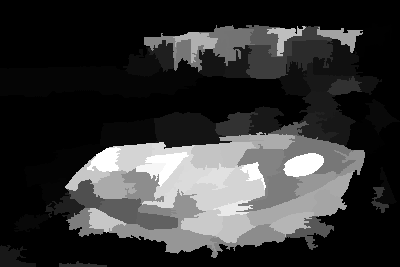} \includegraphics[height=.08\linewidth]{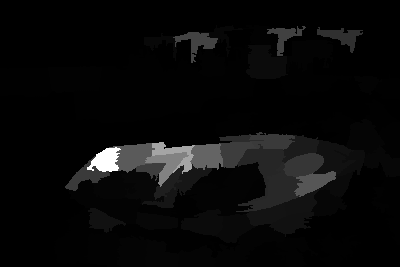}
\includegraphics[height=.08\linewidth]{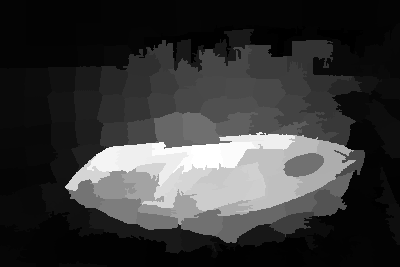} \includegraphics[height=.08\linewidth]{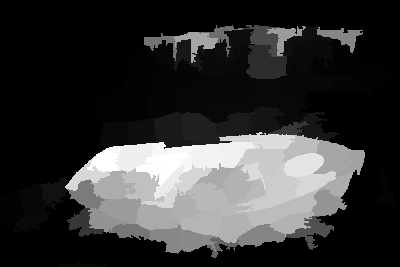} \includegraphics[height=.08\linewidth]{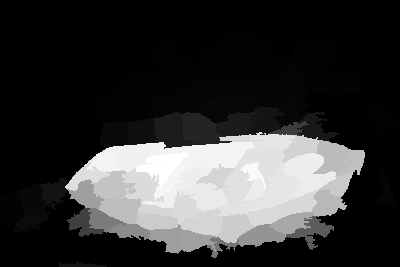}
\includegraphics[height=.08\linewidth]{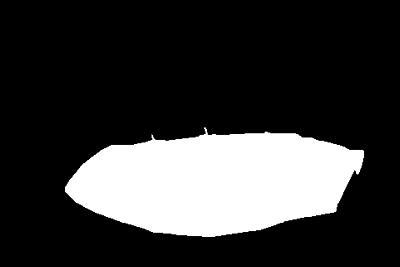}
\\[2pt]
\includegraphics[height=.08\linewidth]{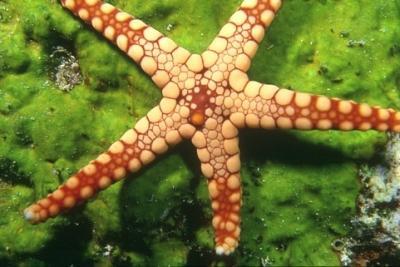}
\includegraphics[height=.08\linewidth]{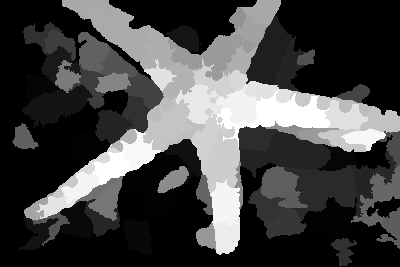} \includegraphics[height=.08\linewidth]{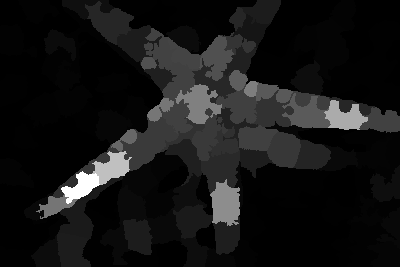}
\includegraphics[height=.08\linewidth]{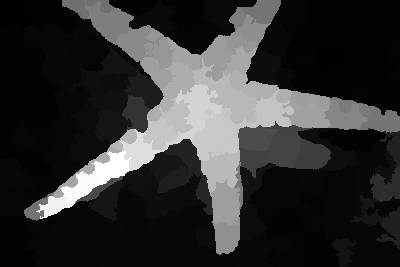} \includegraphics[height=.08\linewidth]{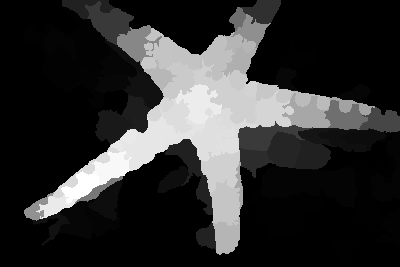} \includegraphics[height=.08\linewidth]{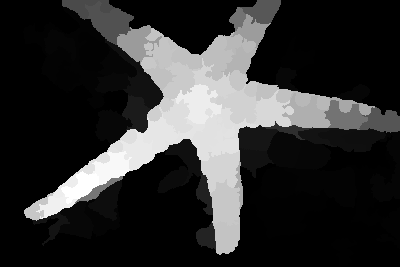}
\includegraphics[height=.08\linewidth]{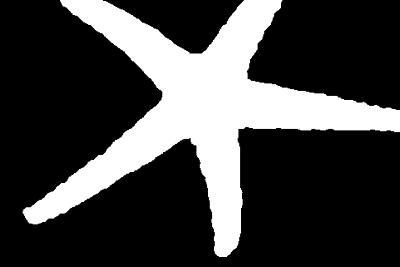}
\\[2pt]
\includegraphics[height=.08\linewidth]{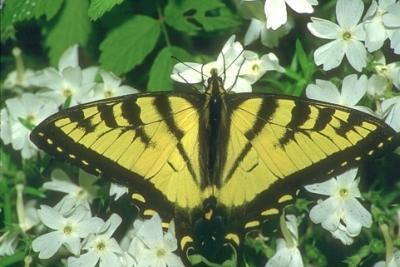}
\includegraphics[height=.08\linewidth]{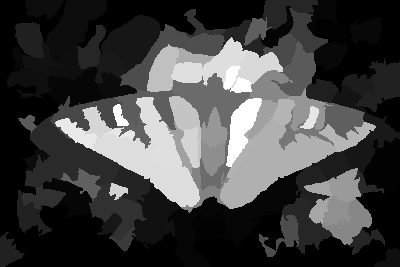} \includegraphics[height=.08\linewidth]{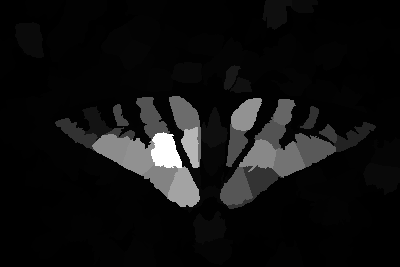}
\includegraphics[height=.08\linewidth]{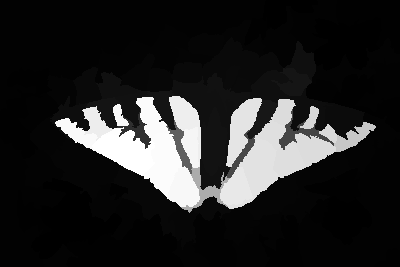} \includegraphics[height=.08\linewidth]{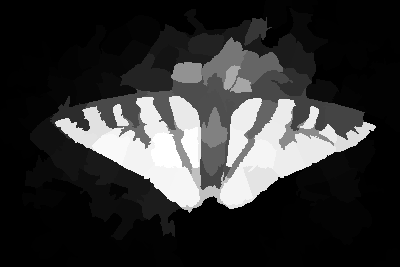} \includegraphics[height=.08\linewidth]{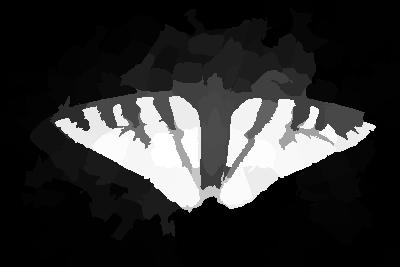}
\includegraphics[height=.08\linewidth]{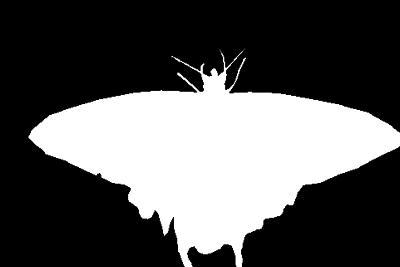}

\centerline{(a)\hspace{.7in}(b)\hspace{.7in}(c)\hspace{.7in}(d)\hspace{.7in}(e)\hspace{.7in}(f)\hspace{.7in}(g)}
\caption{Visual Comparison of Saliency Maps. (a) is the original image. Saliency map obtained using (b) GS \cite{geosal} (c) SF \cite{PerazziKPH12} (d) MR \cite{manifoldranking} (e) SO \cite{salopt} (f) Proposedd and (g) Ground Truth . The proposed approach generates Saliency maps that are accurate, smooth and uniform}
\end{figure*}

\subsection{Foreground Connectivity}
Thresholded Objectness Maps roughly capture the super-pixels which are a part of the foreground. This is exemplified in Figure 2. We propose a novel saliency measure called 'foreground connectivity' that assigns saliency values based on a super-pixels connectivity to the estimated foreground. 
We construct a graph with super-pixels as nodes. Super-pixels that are adjacent in the images are connected by an edge with a weight equivalent to Euclidean distance of their mean LAB values.
We now define foreground connectivity of a super-pixel $R$ as :
\begin{equation}\label{equ:FG}
FG(R) = \frac{\sum^{N}_{k=1} d(R,R_{k}).\delta(R_{k})}{\sum^{N}_{k=1} d(R,R_{k}).(1-\delta(R_{k}))}
\end{equation}
where $d(R,R_{k})$ denotes the shortest distance between $R$ to $R_{k}$  and $\delta(.)$ is 1 for a super-pixel if it is estimated as foreground by the Objectness Map, and $N$ is the total number of super-pixels.

A higher similarity of a super-pixel with the estimated foreground ensures lower value in the numerator and a higher value in the denominator leading to less value of FG (implying higher connectivity). We take the reciprocal of FG and use it as the foreground weights ($w^{fg}$). 

\begin{figure*}
\label{pr}
\centering
\includegraphics[height = 4.5cm, width = 6cm]{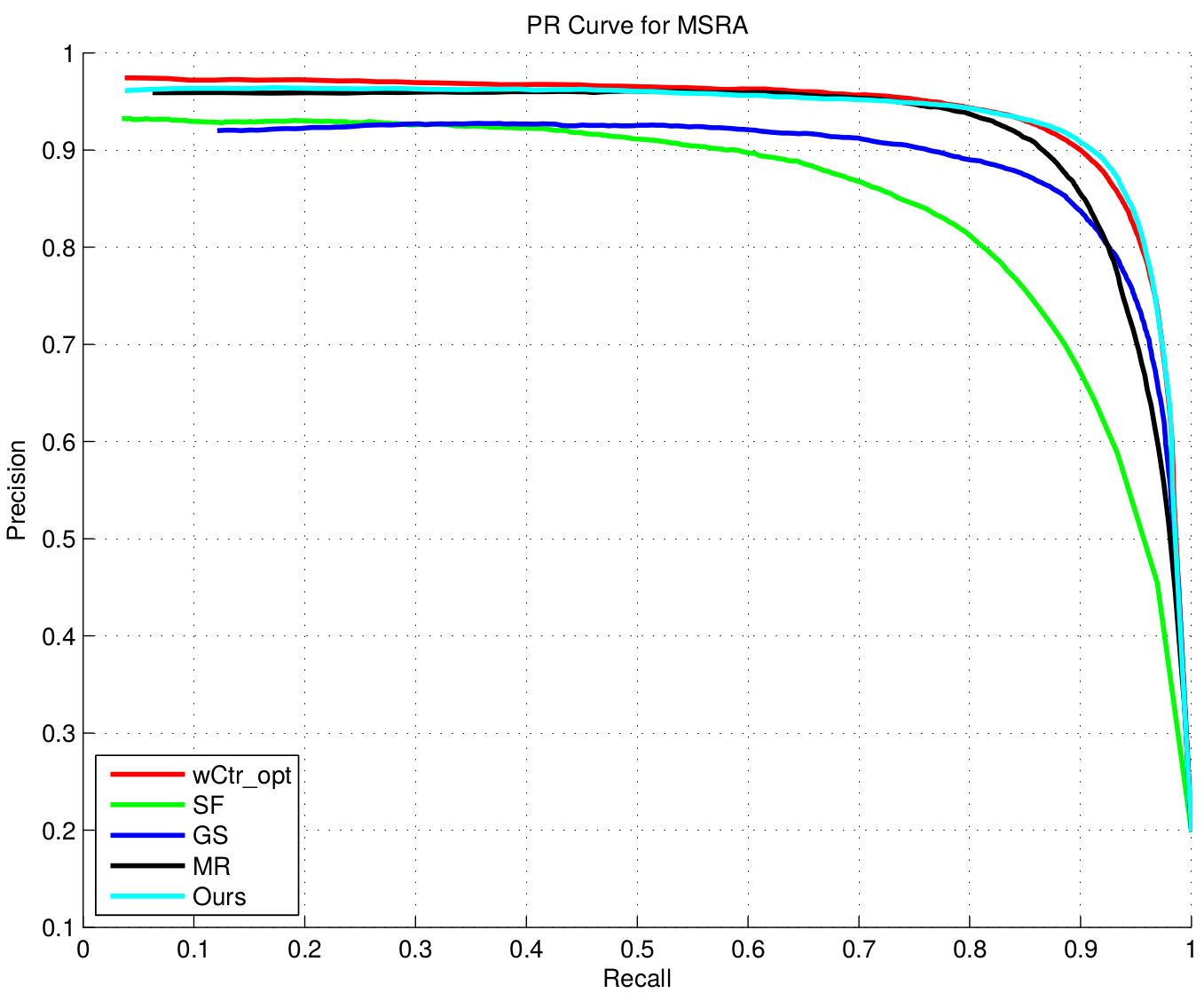}
\includegraphics[height = 4.5cm, width = 6cm]{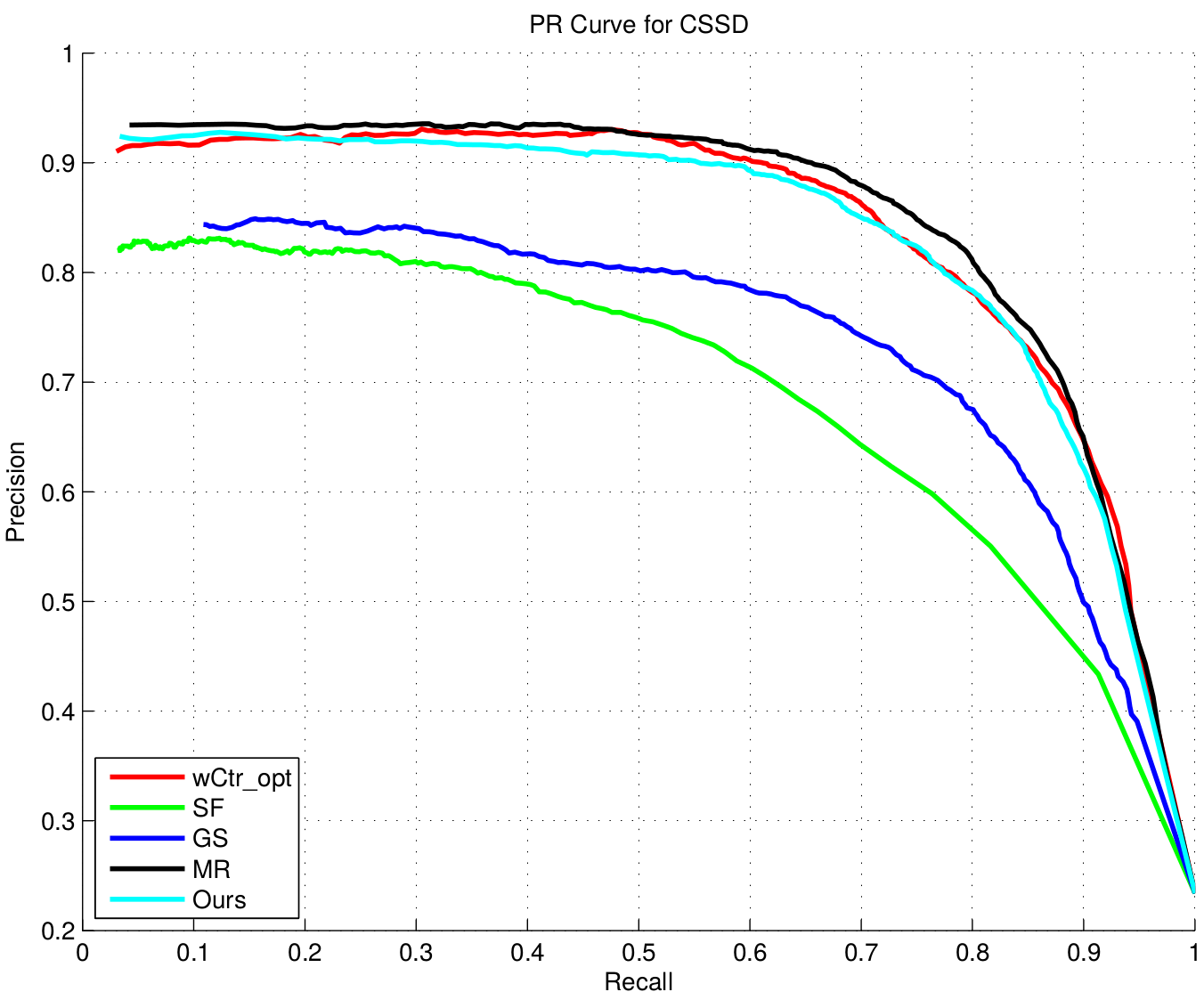}
\centerline{~~~~~(a)~~~~~~~~~~~~~~~~~~~~~~~~~~~~~~~~~~~~~~~~~~~~~~~~~~~~~~~~~~~~~~~~~~~~~~~~~(b)}
\caption{Comparison of PR curves on (a)MSRA-1000 database and (b) CSSD-200 database}
\end{figure*}

\subsection{Saliency Optimization}

Generally, several saliency cues are combined heuristically using weighted summation or multiplication. Instead we use an existing optimization framework to combine our foreground weights with background weights as used in \cite{salopt}. The cost function to be minimized is defined as
\begin{equation}\label{equ:optimization}
\sum^{N}_{i=1} w^{fg}_{i} (t_{i}-1)^{2} + \sum^{N}_{i=1} w^{bg}_{i} (t_{i})^{2} + \sum_{i,j} w_{ij}(t_{i}-t_{j})^{2}
\end{equation}
where $t_i$ denotes the final value of saliency assigned to $p_{i}$ after minimizing the cost, $w^{fg}_{i}$ denotes foreground weights, $w^{bg}_{i}$ denotes background weights associated with super-pixel $p_{i}$. High $w^{fg}_{i}$ encourages $p_{i}$ to take values close to 1 and high $w^{bg}_{i}$ encourages $p_{i}$ to take values close to 0. $w_{ij}$ is the smoothness coefficient.We use the same parameter settings as used in \cite{salopt}.

\section{Results}
We evaluate the proposed approach on two benchmark datasets. The first one is the MSRA-1000 \cite{09cvpr/Achanta_FTSaliency} dataset which is one of the most extensively used databases. The second one is the CSSD dataset \cite{Yan:13}. The MSRA-1000 dataset has large varieties in content and background but are simple and smooth. The CSSD dataset on the other hand has structurally complex images for evaluation. Results obtained on both the datasets are evaluated on ground-truth masks labeled by humans.
We compare the results of our algorithm with recent four state of the art methods : Saliency Filter (SF) \cite{PerazziKPH12} , geodesic Saliency (GS) \cite{geosal} , Manifold Ranking (MR) \cite{manifoldranking} and Saliency Optimization (SO)\cite{salopt}.
Results on MSRA-1000 \cite{09cvpr/Achanta_FTSaliency} and CSSD \cite{Yan:13} are shown in Figure 3.
We evaluate our method using Precision-Recall curves and Mean Absolute Error (MAE).

\subsection{Precision and Recall}
Precision is the fraction of pixels assigned correctly against the total number of pixels assigned salient. Whereas recall is the fraction of pixels labeled correctly in relation to the number of ground truth pixels. Precision and recall vary inversely and hence it is essential to evaluate them simultaneously. Hence we use a Precision-Recall curve similar to previous works . For various value of threshold between $[0 \ldots 255]$, we obtain binary maps and using the ground truth mask,we compute precision and recall values. Figure 4 shows that the proposed approach performs at par with the other state of the art algorithms.
However precision recall curves have some serious limitation. Precision-recall curves do not consider the fraction of pixels correctly assigned as not salient. Presence of pixels incorrectly assigned as salient brings down the performance of saliency map despite it being smooth and having higher values assigned to pixels that are salient.

\subsection{Mean Absolute Error}
To overcome this limitation, we use Mean Absolute Error (MAE) as suggested by \cite{PerazziKPH12}. It measures how similar a saliency map is to the ground truth. For a Saliency map S, ground truth mask G, then MAE is defined as
\begin{equation}\label{equ:MAE}
MAE = \frac{1}{W \times H} \sum_{x=1}^{H}\sum_{y=1}^{W} |S(x,y) - G(x,y)|
\end{equation}
where $H$ and $W$ denote the height and width of the image.
Results have been averaged out over all the images in the database. The proposed approach performs better than the state of the art methods in terms of MAE (see Table 1).

\begin{table}[h]
\centering
\begin{tabular}{ | l | c | r | }

   \hline
   ~ & MSRA & CSSD \\
   \hline \hline
   
  GS \cite{geosal} & 0.109 & 0.178 \\
  SF \cite{PerazziKPH12} & 0.129 & 0.204 \\
  MR \cite{manifoldranking} & 0.085 & 0.150 \\
  SO \cite{salopt} & 0.068 & 0.136 \\
  Proposed & \textbf{0.064} & \textbf{0.132} \\
  \hline
\end{tabular}
		\caption{ Comparison of MAE values of different saliency methods}
\end{table}

\subsection{Running Time}
The average running time of the proposed approach on an Intel Core i5-4200U CPU @ 1.60 GHz with 6GB RAM is 0.27s excluding pre-processing and Superpixel segmentation using SLIC \cite{slic}.

\section{Conclusion}

In this paper, we present a simple and efficient method that utilizes objectness Proposals and foreground connectivity measure to detect salient objects in an image. Unlike recent methods, we obtain our saliency maps by estimating foreground regions in an image instead of using boundary priors. Our method combined with the optimization framework produces accurate and smooth saliency maps that perform better than other methods in terms of MAE when tested on two widely used datasets.
In future, we plan to investigate better cues rather than depending on contrast or boundary prior alone, better connectivity measures and better objectness proposal techniques that can perform well with backgrounds that are even more complex.

\newpage

\bibliographystyle{IEEEbib}
\bibliography{strings,refs}

\end{document}